\documentclass[letterpaper, 10 pt, journal, twoside]{ieeetran} 

\IEEEoverridecommandlockouts                              

\usepackage[english]{babel}

\usepackage{graphicx}
\usepackage{animate}
\usepackage{epsfig} 				
\usepackage{epstopdf}

\usepackage{listings}
\usepackage{color}
\usepackage{nameref}
\usepackage{hyperref}
\usepackage[colorinlistoftodos]{todonotes}
\usepackage{amsmath}	 			
\usepackage{amssymb}  				
\usepackage{dsfont}			
\usepackage{mathtools}

\usepackage{epigraph}
\usepackage{lscape}
\usepackage[]{nomencl}				
\usepackage{algorithm}
\usepackage{algorithmic}
\usepackage{multicol}
\usepackage{multirow}
\usepackage{etoolbox}

\usepackage{caption}
\usepackage{subcaption}

\usepackage{wrapfig}
\usepackage{multimedia}

\usepackage{siunitx}
\usepackage{flushend}

\usepackage{floatflt}

\usepackage{url}

\newcommand{\citet}[1]{\cite{#1}}
\newcommand{\citep}[1]{\cite{#1}}

\newcommand{\email}[1]{\href{mailto:#1}{\nolinkurl{#1}}}

\newcommand{\link}[1]{\colora{\url{#1}}}

\renewcommand{\sec}[1]{Section~\ref{#1}}
\newcommand{\fig}[1]{Fig.~\ref{#1}}

\newcommand{\tab}[1]{Table~\ref{#1}}

\usepackage{bm}
\newcommand{\sensor}[0]{DIGIT}

\newcommand{\repo}[0]{\url{www.digit.ml}}

\newcommand{\raleditor}[0]{Hong Liu}

\title{DIGIT: A Novel Design for a Low-Cost Compact High-Resolution Tactile Sensor with Application to In-Hand Manipulation}

\author{Mike Lambeta$^{\ast}$, Po-Wei Chou$^{\ast}$, Stephen Tian$^{\ast}$, Brian Yang$^{\ast}$, Benjamin Maloon, Victoria Rose Most, Dave Stroud, Raymond Santos, Ahmad Byagowi, Gregg Kammerer, Dinesh Jayaraman, and Roberto Calandra%
\thanks{Manuscript received: September, 10, 2019; Revised December, 2, 2019;
Accepted January, 27, 2020.}
\thanks{This paper was recommended for publication by Editor \raleditor{} upon
evaluation of the Associate Editor and Reviewers' comments.}%
\thanks{$^\ast$Equal contribution.}
\thanks{All authors are with Facebook, Menlo Park, CA, USA\newline
{\tt\small \{lambetam, poweic, stephentian, brianhyang, benjamin.maloon, victoriamost, dstroud, raysantos, abyagowi, greggk, dineshj, rcalandra\}@fb.com}}%
\thanks{Digital Object Identifier (DOI): 10.1109/LRA.2020.2977257}
}

\begin{document}

\maketitle


\begin{abstract}
	Despite decades of research, general purpose in-hand manipulation remains one of the unsolved challenges of robotics. One of the contributing factors that limit current robotic manipulation systems is the difficulty of precisely sensing contact forces -- sensing and reasoning about contact forces are crucial to accurately control interactions with the environment. As a step towards enabling better robotic manipulation, we introduce DIGIT, an inexpensive, compact, and high-resolution tactile sensor geared towards in-hand manipulation. DIGIT improves upon past vision-based tactile sensors by miniaturizing the form factor to be mountable on multi-fingered hands, and by providing several design improvements that result in an easier, more repeatable manufacturing process, and enhanced reliability. We demonstrate the capabilities of the DIGIT sensor by training deep neural network model-based controllers to manipulate glass marbles in-hand with a multi-finger robotic hand. 
To provide the robotic community access to reliable and low-cost tactile sensors, we open-source the DIGIT design at \repo{}.
\end{abstract}



\section{INTRODUCTION}
	
\IEEEPARstart{R}{obots} are not yet capable of achieving the same level of manipulation dexterity as humans.
One contributing factor is the difficulty of precisely estimating contact forces.
Forces are an important representation to understand and plan interactions with the environment -- grasping a small screw, inserting a key, and manipulating a glass marble are all examples that highlight the need for accurate control of contact forces. 
Touch is a crucial sensory modality for both humans~\citep{Johansson2009Coding} and robots~\citep{Calandra2017Feeling}, as it provides a natural, direct, and virtually noiseless way to measure forces -- unlike any other sensor modality.
In recent years, the use of touch sensing has became a relevant topic in the robotic community, and a large body of literature studies how to integrate touch to improve perception and manipulation~\citep{Bekiroglu2012,Cockbum2017,Veiga2018Hand,Lee2019Making}.
Despite the existence of many different types of tactile sensors~\citep{Ueda2005Development,Yousef2011Tactile,Fishel2012Sensing,Yamaguchi2016Combining,Yuan2017GelSight,Donlon2018GelSlim,Ward-Cherrier2018tactip,Church2019Tactile,Sferrazza2019Design}, the main bottleneck for wide adoption of touch sensing in robotic manipulation is the lack of sensors that fulfill at the same time all the requirements of being 1) high resolution, 2) highly sensitive, 3) reliable, 4) easy to use, 5) compact, and 6) inexpensive. 

To better fulfill these requirements, in this paper, we present the design of a novel tactile sensor.
Our new sensor, ``\sensor{}'', introduces several critical improvements over past vision-based tactile sensors: a smaller form factor to enable in-hand manipulation on multi-finger hands, a streamlined manufacturing process that reduces cost and assembly time and potentially enables large-scale manufacturing, and enhanced mechanical reliability that substantially extends its lifespan.
In addition, \sensor{} retains the rich and sensitive measurements characteristic of previous vision-based sensors~\citep{Johnson2009Retrographic,Yamaguchi2016Combining,Yuan2017GelSight,Donlon2018GelSlim}. Moreover, \sensor{} is designed to be modular so that individual components may be replaced easily, and comes with a software interface that facilitates ``plug-and-play'' usage.

\begin{figure}[t]
    \centering
    \includegraphics[width=\linewidth]{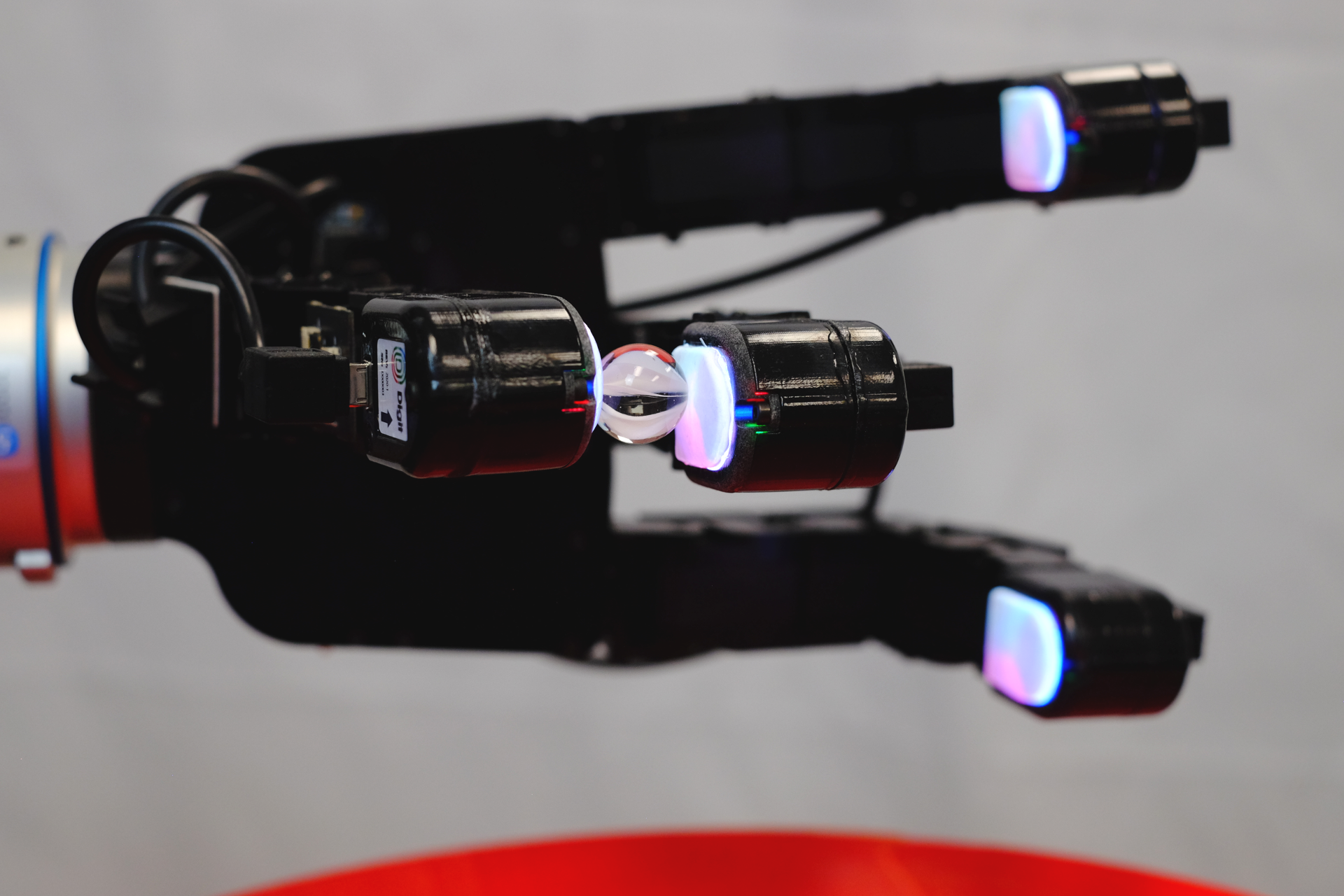}
    \caption{\sensor{}s mounted on an Allegro multi-finger hand. To validate our sensor design, we learn to manipulate glass marbles between two fingers.}
    \label{fig:teaser}
\end{figure}
%
The contribution of this paper is two-fold. 
First, we present the design and manufacturing process of \sensor{}, and analyze the properties of the resulting sensor.
Second, we demonstrate the sensor by learning to manipulate small objects with a multi-finger hand from raw tactile inputs. 
The learning approach used is based on tactile-MPC~\citep{Tian2019Manipulation}. 
However, while tactile-MPC has thus far been demonstrated on a single touch sensor, we are interested in handling multiple touch sensors from different fingers.
To scale up tactile-MPC, we propose new approaches for dynamics model learning and task specification that dramatically reduce the computational cost.

\sensor{} aims to stimulate future research in tactile sensing within the robotics community, by providing an affordable, robust, and easy-to-use tactile sensing platform that effectively removes the complexity typically associated with tactile sensors.
For this reason, in conjunction with this paper, we release the design of the sensor at \repo{}.

	

\section{RELATED WORK}
\label{sec:related}

The design of tactile sensors has been an active field of research for many years~\citep{Dahiya2010Tactile} with a large range of technologies used for measuring forces~\citep{Cannata2008embedded,Fishel2012Sensing,Le2017highly}.

A class of sensors that has recently proved popular and versatile in the robotic community, are the vision-based tactile sensors.
The idea at the base of vision-based sensors is to measure contact forces as changes in images recorded by a camera, typically through the use of a deformable elastomer~\citep{Begej1988Planar,Hristu2000performance,Kamiyama2004Evaluation,Ueda2005Development,Johnson2009Retrographic}. 
Compared to other classes of tactile sensors, vision-based sensors often provide advantages in terms of spatial resolution, higher sensitivity, and manufacturing cost, although resulting in bulkier form factors.
Previous vision-based tactile sensors include TacTip~\citep{Ward-Cherrier2018tactip,Church2019Tactile}, FingerVision~\citep{Yamaguchi2016Combining}, GelSight~\citep{Yuan2017GelSight,Donlon2018GelSlim} and several other~\citep{Sferrazza2019Design,Shimonomura2019Tactile}.
Among these, GelSight sensors have been quite popular in the recent robotic literature, and several sensors have been presented~\citep{Yuan2017GelSight,Donlon2018GelSlim} that use a soft reflective elastomer with printed markers as the contact surface, and outputs images of the surface deformations. 
\sensor{} improves over existing GelSight sensors in several ways: by providing a more compact form factor that can be used on multi-finger hands, improving the durability of the elastomer gel, and making design changes that facilitate large-scale, repeatable production of the sensor hardware to facilitate tactile sensing research.
FingerVision~\citep{Yamaguchi2016Combining} proposed the use of a \emph{transparent} elastomer with markers, thus allowing the camera to be used also for seeing the objects during the approach phase.
The main disadvantage of this design is the decrease in tactile resolution, as now only the movement of the markers in the elastomer provides touch information.
Similar limitation applies to the TacTip sensors~\citep{Ward-Cherrier2018tactip,Church2019Tactile} which measure deformation of the elastomer through the movements of physical internal pins.
While \sensor{} is by default equipped with reflective elastomers, its modular design makes it easy to swap in a FingerVision-style transparent elastomer, or a TacTip-style elastomer with markers, as discussed in \sec{sec:sensor}.
For a more complete review of vision-based tactile sensors, we point the readers to \citet{Shimonomura2019Tactile}.

The integration of tactile sensing for robotic manipulation has long been a research focus~\citep{Yousef2011Tactile,Kappassov2015Tactile}. 
A key bottleneck in prior efforts towards tactile manipulation is that it is often difficult to extract and integrate meaningful features from high-resolution tactile sensors in control algorithms.
Due to this, much of prior work in learning manipulation relies purely on vision or proprioception.
%
Recent work on in-hand manipulation \citep{OpenAI2018learning} includes the use of model-free reinforcement learning to learn in-hand object reorientation. However, this method requires a careful estimation of robot state, which necessitates many tracking cameras for each of the fingers of the hand. This setup can be physically restrictive of the types of settings the system may operate in. 
Deep reinforcement learning has also been applied to learn a variety of dexterous manipulation skills using low-cost robotic hands~\citep{Zhu2019Dexterous}. 
In this work, we focus on learning dexterous in-hand tasks requiring delicate control, which necessitates the use of tactile sensing for precise feedback. Our sensors are compact and attached directly to the robot end-effector, and thus applicable in many real-world scenarios. 

Learning approaches are particularly suitable to integrate high-dimensional information from vision-based tactile sensors, allowing us to take advantage of valuable touch feedback during control. 
Touch sensing has shown promise in learning methods for in-hand manipulation, accomplishing a rolling task of a large object between two fingers of a robotic hand using MEMS barometers as tactile feedback~\citep{van2015learning}
The GelSight sensor in particular has also shown success in learning models to predict grasp success of complex and varying geometries~\citep{Calandra2018More}. 
Model learning methods have been used to solve basic under-actuated manipulation tasks with vision-based tactile sensors~\citep{Tian2019Manipulation} as well as the BioTac~\citep{Veiga2017Tactile}. 
Due to hardware limitations, many more challenging tasks which approach the abilities of human in-hand manipulation remain unexplored.
We build off these model learning methods to tackle more complex manipulation tasks using the  improvements afforded by our \sensor{} sensors.


\section{\sensor{}: A LOW COST, COMPACT, HIGH-RESOLUTION TACTILE SENSOR}
\label{sec:sensor}

\begin{figure}[t]
    \centering
    \includegraphics[width=.98\linewidth]{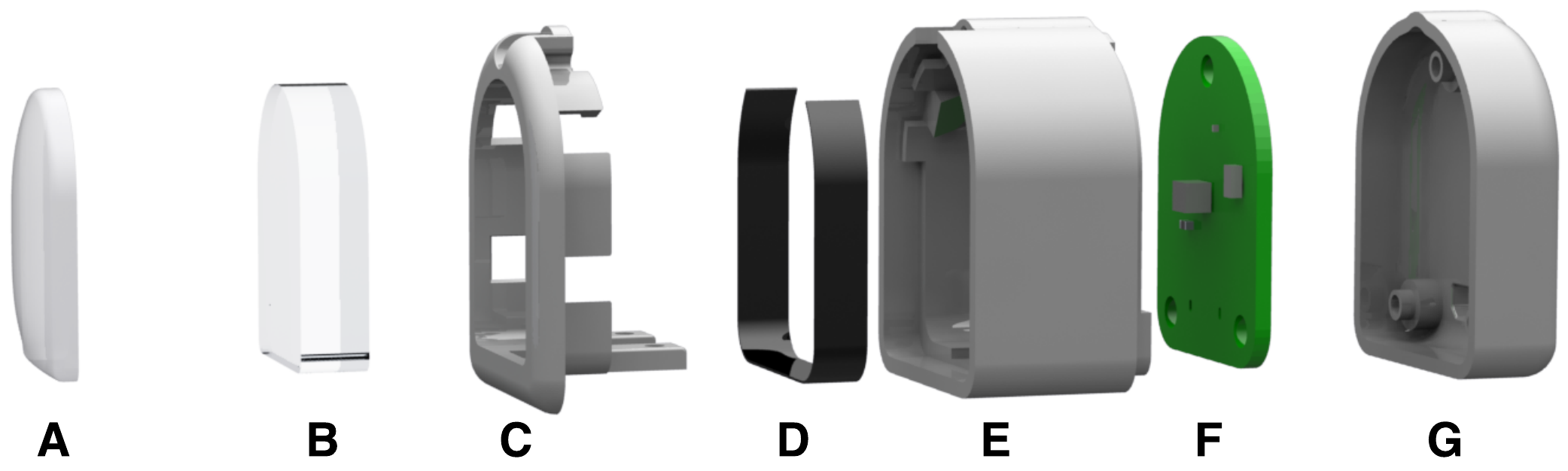}
    \caption{Exploded view of a single \sensor{} sensor. A) elastomer, B) acrylic window, C) snap-fit holder, D) lighting PCB, E) plastic housing, F) camera PCB, G) back housing.}
    \label{fig:exploded}
\end{figure}
\begin{table}[tb]
\centering
\caption{Comparison of \sensor{}, GelSight, and GelSlim.\\ $^*$ Considering the manufacturing of 1000 pieces}
\resizebox{\linewidth}{!}{%
\begin{tabular}{l|c|>{\centering}p{1.8cm}|c|}
& \sensor{} (Ours) & Fingertip GelSight~\citep{Yuan2017GelSight} & GelSlim~\citep{Donlon2018GelSlim} \\ \hline
Size [mm] & 20x27x18 & 35x60x35 & 50x205x20\\
Weight [g] & 20 & NA & NA\\
Sensing field [mm] & 19x16 & 18x14 & 30x40\\
Image Resolution & 640x480 & 1920x1080 & 640x480\\
Image FPS & 60 & 30 & 60\\
Cost components [\$]& 15$^*$ & $\sim$30 & NA\\
\end{tabular}
}
\label{tab:sensor}
\end{table}

\begin{figure*}[t]
    \centering
    \includegraphics[width=\linewidth]{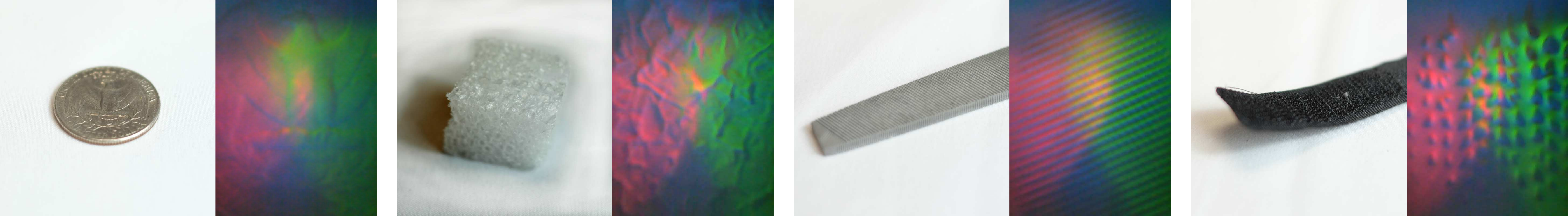}
    \caption{Object under test and corresponding raw measurements taken using \sensor{}. The measurements taken from \sensor{} clearly capture sub-millimeters structures.}
    \label{fig:impressions}
\end{figure*}

We now present \sensor{}, our new vision-based tactile sensor. 
While previous vision-based tactile sensors offer unparalleled high spatial resolution raw tactile sensing, they have three main limitations compared to other tactile sensors: (i) they have relatively bulky form factors, (ii) the use of soft materials at the surface of contact makes them susceptible to wear out quickly compared to other sensors, and (iii) they require a complex (largely manual) manufacturing process that leads to high variability between sensors, so that replacing a damaged sensor is not easy -- the system might have to be re-calibrated or retrained to adapt to the characteristics of the new sensor.

\sensor{} inherits the advantages of vision-based tactile sensors, while also addressing these three drawbacks. 
First, \sensor{} is designed to be sufficiently physically compact to fit on an array of end effectors or multi-fingered robot arms, as seen from \fig{fig:teaser}. 
Second, \sensor{}'s gel is designed to be more robust and at the same time more easily interchangeable than previous designs, resulting in an overall more rugged sensor. 
Finally, the design of \sensor{} incorporates new automated manufacturing techniques, emphasizing tool-less assembly and commercial off-the-shelf components to permit rapid large-scale, repeatable manufacture at very low costs. 
\sensor{}'s total estimated manufacturing cost is approximately 15 USD per sensor (PCB: 1.5 USD, electronic components: 8 USD, plastics: 2 USD, gel: 3 USD), when manufactured in a batch of 1000. 
In \tab{tab:sensor}, we compare DIGIT against two popular vision-based tactile sensors.
In the following sections, we overview the design decisions through which \sensor{} achieves these advantages.

\subsection{Mechanical Design}
\begin{figure}[t]
    \centering
    \includegraphics[height=2.7cm]{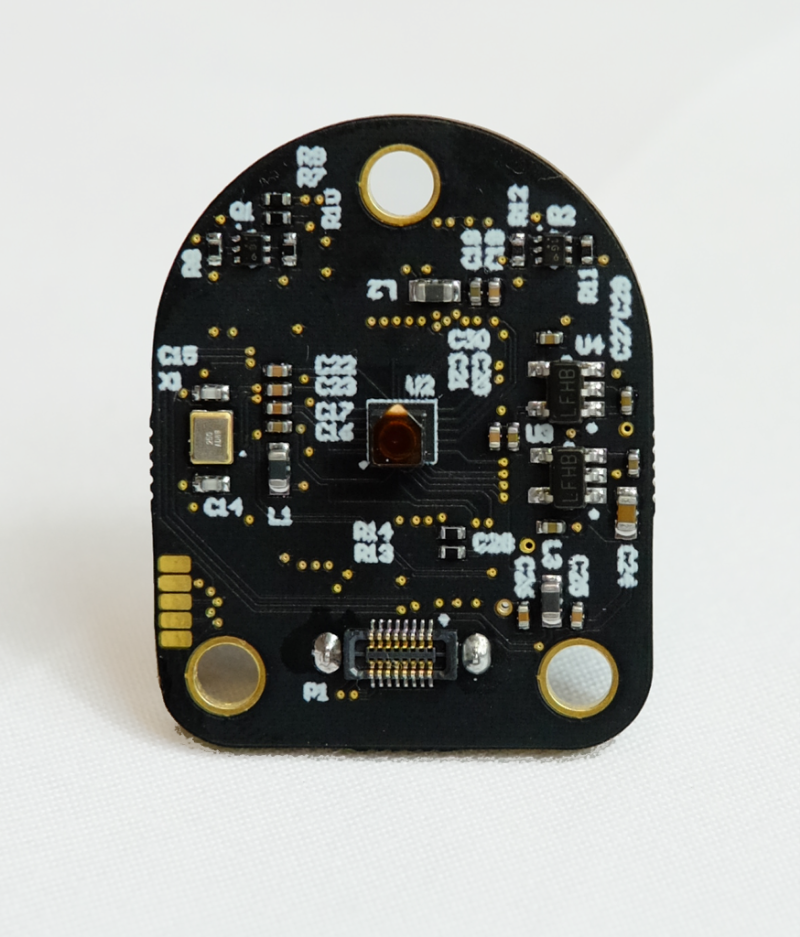} 
    \hfill 
    \includegraphics[height=2.7cm]{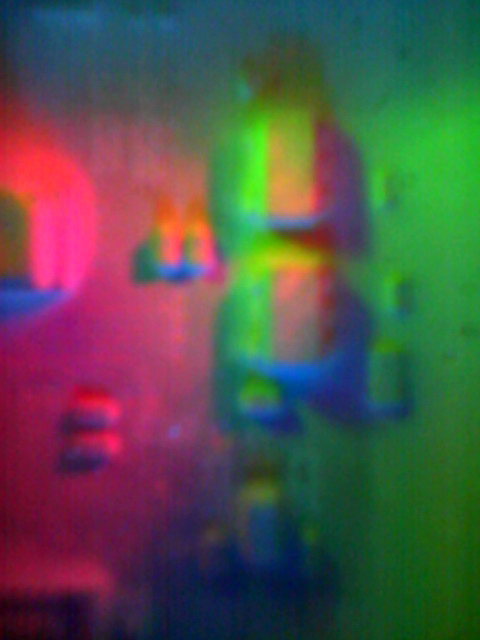}
    \hfill    
    \includegraphics[height=2.7cm]{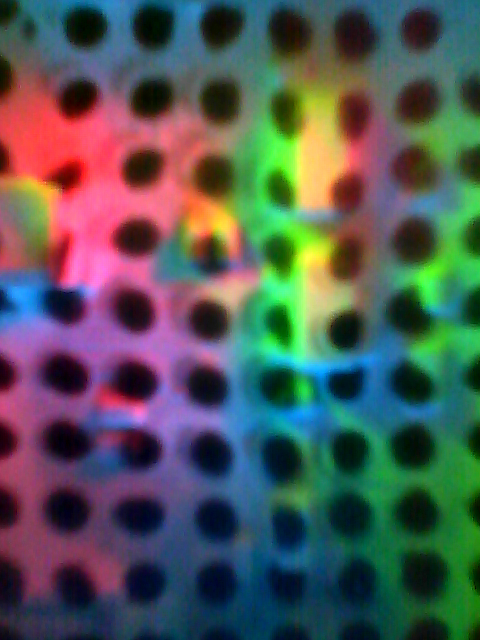}
    \hfill
    \includegraphics[height=2.7cm]{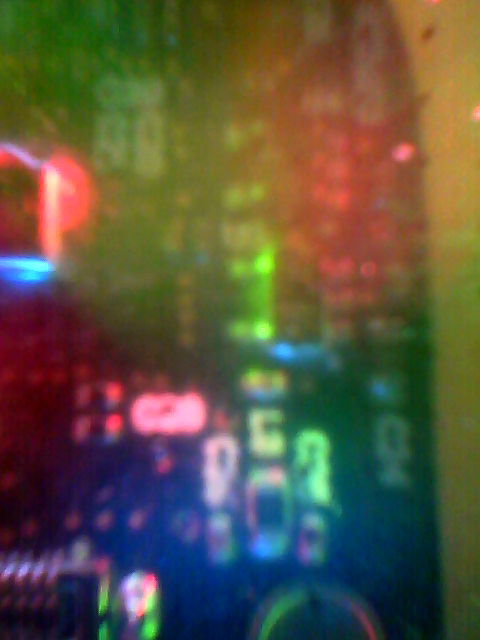}
    \caption{\sensor{} supports different types of elastomers which can be rapidly replaced thanks to its mechanical design. Here we show readings when touching an object (\textit{left}) using three different elastomers: reflective, reflective with markers, and transparent with markers. Each elastomer can have different benefits for different applications, e.g., reflective to measure textures, reflective with markers to compute optical flow.}
    \label{fig:geltypes}
\end{figure}

An exploded view of the mechanical design of \sensor{} is presented in \fig{fig:exploded}. 
A full \sensor{} has dimensions \SI{20}{\milli\meter} width x \SI{27}{\milli\meter} height x \SI{18}{\milli\meter} depth, and weighs approximately \SI{20}{\gram}. 
\sensor{} has a plastic multi-body three-piece enclosure that is easy to 3D print for prototyping, or injection mold for large-scale production. The camera and gel are mounted to this body using ``press fit'' connections so that any one component may be easily swapped out upon breakage or wear and tear. 
Additionally, the plastic housing can be swapped to allow for different focal lengths, and the elastomer can be easily replaced through a single screw.
For example, it is possible to swap in task-specific elastomers into the same \sensor{} unit, with hardness and opaqueness tuned to the required sensitivity and expected forces in that task.
Examples shown in \fig{fig:geltypes} are purely reflective elastomers to accurately measure surface and texture~\citep{Johnson2009Retrographic}, reflective elastomers with markers to compute optical flow~\citep{Yuan2017GelSight}, and transparent elastomers with markers to control finger position during grasping~\citep{Yamaguchi2016Combining}. 
The multi-body design of \sensor{} also significantly simplifies the assembly process and makes it easy to scale repeatably.

\subsection{Electronic Design} 
Instead of relying on existing camera solutions, we decided to custom-design the electronics that control camera characteristics, illumination, and video capture.
By doing so, the resulting electronics fit within an area of \SI{7}{\centi\meter\squared}, only slightly larger than a human fingertip.
For the camera, we use an Omnivision OVM7692, a \SI{60}{fps} color CMOS hosting a microlens array with focal length \SI{1.15}{\milli\meter} and depth of field \SI{30}{\centi\meter}. 
A custom PCB connects the camera to a SuperSpeed USB 3.0 hub to facilitate connecting multiple \sensor{}s to a single USB port on a host computer. 
This PCB also allows manual control of illumination intensity for the three RGB LEDs, which can provide a maximum of \SI{4} lumens over the elastomer surface.

\subsection{Elastomer Design}

Vision-based tactile sensors rely on soft deformable elastomeric materials at the surface of contact, which often leads to significant wear and tear, altering the characteristics of the sensor over repeated use. 
GelSight gels contain a transparent base layer beneath an opaque image transfer layer at the surface of contact. 
The image transfer layer's deformations, observed by the camera, constitute the tactile percepts from the sensor. 
This layer is in contact with objects during use, and thus liable to suffer wear and tear over time.

We have developed a new gel manufacture process that increases the lifetime and reliability of the gel, while making it amenable to large scale production, all without compromising tactile sensing performance.
For \sensor{}, we construct the elastomer in three stages. 
A silicone-based white pigment is added to the mold with an airbrush and left to cure with a chemical kicker to produce an image transfer layer with controlled uniform thickness. 
The base layer silicone is then applied to the finger-like shape mold and left to cure. Later, the silicone is removed from the mold, and glued onto an acrylic window using Smooth-On Sil-Poxy, an optically clear silicone adhesive. 
This acrylic-gel unit can then be press-fit into the body during assembly, as described above. 
For the silicone, we use Smooth-On Solaris, a type of silicone typically used to coat photo-voltaic cells.

A thick image transfer layer deforms less, resulting in loss of spatial resolution in the tactile sensing outputs. On the other hand, thin layers are more prone to damage. In the design process we iterated over the thickness of the image-transfer layer, trading off ruggedness and sensitivity. 
\fig{fig:impressions} shows tactile sensing outputs from \sensor{} when contacting various objects, demonstrating its sensitivity. 
Future iterations of \sensor{} could include gels with different thicknesses based on the operating range of forces that we would like the sensor to be most sensitive to. 
Next, we evaluate the robustness of our elastomer.

\subsection{Mechanical Robustness of the Elastomer}
\label{sec:robustness} 
\begin{figure}[t]
  \centering
    \includegraphics[width=\linewidth]{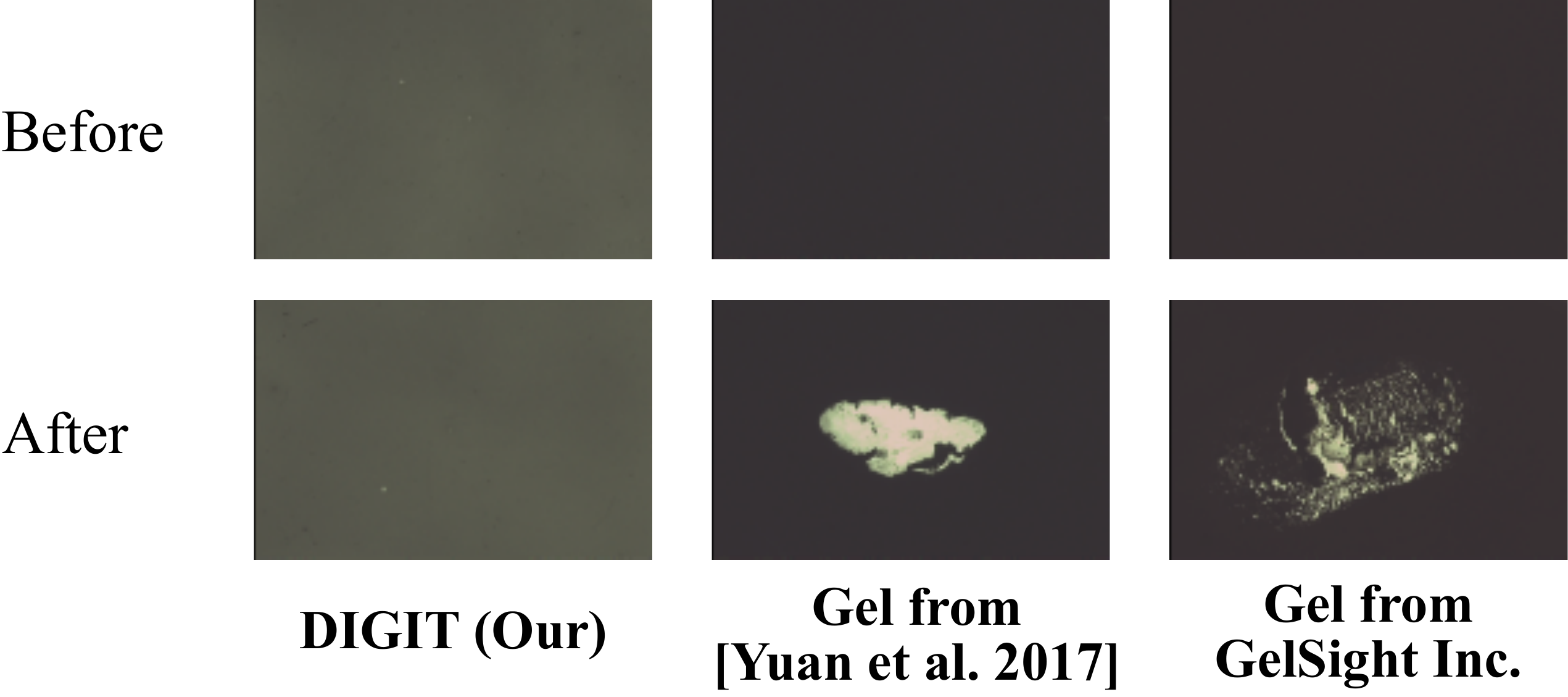}
  \caption{Surface of the different gels after 5 abrasion passes. In both the gel from~\citep{Yuan2017GelSight} and the one from GelSight Inc. the coating layer is visibly damaged.}
  \label{fig:abrasion}
\end{figure}

We tested the mechanical characteristics of the \sensor{} elastomer against a gel provided by Yuan et al~\citet{Yuan2017GelSight}, and one provided by GelSight Inc. The gel provided by GelSight Inc. was not designed for robotic applications, but rather for high-resolution 3D measurements and uses a very thin and opaque coating layer.
To perform this test, we used an industry-standard linear abrasion device with a total calibrated weight of \SI{1.7}{\newton} and a H-18 Calibrade medium abrasive plunger tip. 
For each gel, we performed a cycle of 5 linear sweeps across the surface of the gel. 
After each cycle, the gel was illuminated from below and optically observed for light transmittance via tears or discontinuities on the gel surface, indicating loss of the opaque image transfer layer. 
\begin{table}[t]
\centering
\caption{Reliability comparison of various gels via abrasion testing. Degradation is measured as percentage of increase in transmittance. Higher values indicate that the coating of the elastomer is wearing out more. Our elastomer demonstrates the lowest amount of degradation over time compared to the other gels evaluated.}
\begin{tabular}{l|ccc}
 &  \multicolumn{3}{l}{\textbf{Degradation [\%]}} \\
\multicolumn{1}{l|}{\textbf{Gel / Abrasion Passes}} &  5 & 10 & 15 \\ \hline
\sensor{} (Ours) & 0 & 0.3 & 0.3 \\
Gel from~\citep{Yuan2017GelSight} &  276 & 482 & 805 \\
Gel from GelSight Inc. & 475 & 662 & 918
\end{tabular}
\label{tab:abrasion}
\end{table}
We record the luminous flux per unit area of the transmitted light through a light meter after 5, 10, and 15 abrasion cycles. 
\tab{tab:abrasion} shows the increase in flux over time as the gels deteriorate under abrasion. 
The initial absolute luminous flux was significantly different for the three gels at the beginning of testing, with the \sensor{} gel transmits 676 Lux out of 1255 Lux, while the other two gels only transmit 17 and 16 Lux respectively. 
While \sensor{}'s gel exhibits a higher degree of translucency, the object impressions in \fig{fig:impressions} and our in-hand manipulation results in \sec{sec:result} demonstrate that this does not adversely affect tactile sensing performance.
\fig{fig:abrasion} shows the three gels before and after a single cycle of 5 passes, showing clearly how the \sensor{}'s gel is nearly unaffected under abrasion, while both other gels suffer significant damage. 
The damage sustained on the other two gels rendered them unusable due to large tears and removal of surface material. 

%


\section{LEARNING IN-HAND MANIPULATION WITH HIGH-RESOLUTION TACTILE SENSORS} 
\label{sec:approach}

	Fine-grained in-hand manipulation is a longstanding task in robotics that has been bottlenecked by the absence of appropriate tactile sensors. Prior works have shown that high-resolution tactile sensing enables fine tactile control tasks~\cite{Calandra2018More,Tian2019Manipulation}, but those sensors were too bulky to demonstrate in-hand manipulation with standard-sized robotic hands. 
As explained above, \sensor{} is much more compact and fits comfortably onto an Allegro robotic hand. 
This means that, for the first time, it is possible to equip a robotic hand with high-resolution camera-based tactile sensing on all fingers.
This in turn opens up new possibilities for tactile sensing-enabled fine in-hand manipulation. 
As a demonstration, we use \sensor{}s on an Allegro hand to teach it to hold and manipulate a marble within a precision grip between the thumb and the middle finger equipped with \sensor{}s and move the marble to the desired goal locations.
This tactile control task is significantly more complex than others that have been demonstrated before~\cite{Calandra2018More,Tian2019Manipulation}.

\begin{figure*}[t]
    \centering
    \includegraphics[width=\linewidth]{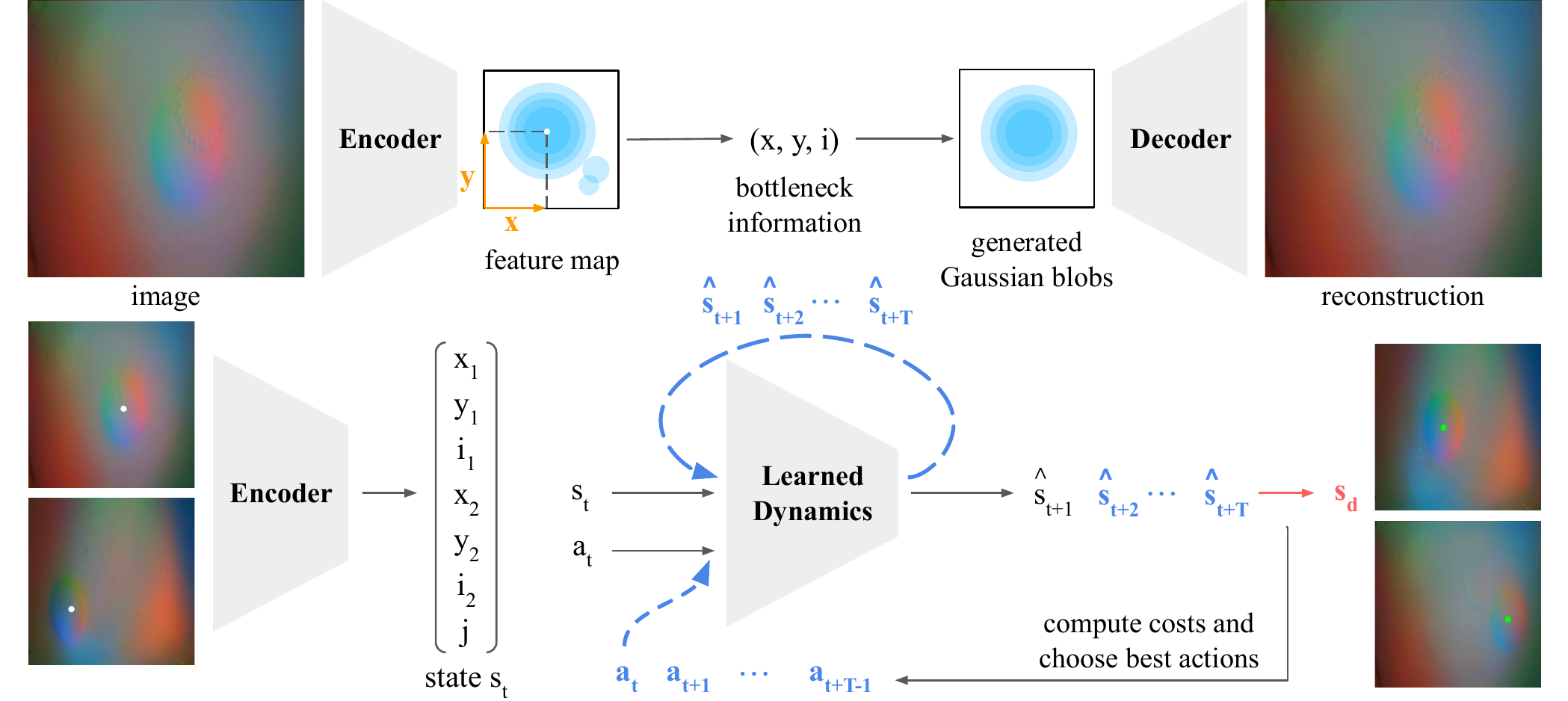}
    \caption{
        System diagram of the self-supervised marble detector (top) and model predictive control using the learned dynamics for marble manipulation (bottom). We first used the encoder part of the autoencoder network to detect the position of the marble from the tactile observations. We then trained a forward dynamics model predicting the position of the marble at the next time step, which we subsequently used to perform model predictive control. 
        At each time step, an optimizer is used to find the best sequences of actions $\mathbf{a}^{*}_{t:t+T-1}$ that moves the marble from the current position to the specified target position~$\mathbf{s}_d$, and the first action $\mathbf{a}^{*}_t$ is applied to the robot.
    }
    \label{fig:mpc-diagram}
\end{figure*}

\subsection{Task Setup}
We mount the Allegro hand on a Sawyer robotic arm, as seen in \fig{fig:teaser}.
At the beginning of each trial, a marble is raised by a metallic stand, similar to a golf tee, mounted on a linear motor, and the arm executes a preprogrammed motion to pick up the marble from this platform with a pincer grip, between the thumb and another finger.  
It must learn to roll its fingers carefully over the marble to manipulate it to the desired configuration. 
This requires modeling the slipping and rolling dynamics of the marble over the small, curved and deformable \sensor{} surfaces under various degrees of pressure from both fingers, an extremely challenging task.

\subsection{Self-supervised Data Collection}
For learning the dynamics model, we collected data from 4800 trials, of which we set aside 950 for validation.
In each trial, after picking up the marble, we move the fingers randomly over the marble for approximately 10 seconds by issuing 20 angular displacement commands to four joint servos along each finger, for an 8-D action space. We recorded videos from both \sensor{}s, the joint angular positions of eight servos (denoted as $j$) and the joint angular displacement commands issued to them (denoted as action $a$). 
At the end each trial, the marble is dropped into a bowl, at the bottom of which is the metallic platform that raise the marble again for the next trial. 
This automatic ``reset'' mechanism permits collecting data from thousands of trials autonomously, without any human intervention.

\subsection{Tactile Predictive Model} 
Tian et al~\citet{Tian2019Manipulation} applied a visual predictive model to model the dynamics of an optical tactile sensor's observations under 3-D end-effector position changes for tactile control. 
We have a more complex setup involving two tactile sensors on the two fingers, and our control commands are 8-D angular displacements corresponding to the eight servos composing these fingers. 
To handle this increased complexity, we use a different choice of predictive model, based on the ``Structural VRNN'' architecture proposed in \cite{Minderer2019Unsupervised}, which is also closely related to approaches proposed in~\cite{zhang2018unsupervised,kulkarni2019unsupervised}.

We first train an autoencoder with a structural bottleneck that learns to detect keypoints of the object representing the factors of variation in the input data, so that modeling the dynamics of those keypoints suffices to perform video prediction. 
The autoencoder consists of a keypoint encoder and a decoder, and we used a tiny version of ResNet-18 as the backbone network for both of them. 
he encoder processes the input image and outputs $K$ feature maps. 
From each of the $K$ feature maps, we obtain a ``keypoint'' prediction $k = [x, y, i]$ consisting of the 2D location $x, y$ that has maximum activation and also an ``intensity'' scalar $i$ representing the average magnitude of the activation. 
At decoding time, for each one of the $K$ keypoint predictions, we draw a Gaussian blob on an empty feature map. 
Then, the decoder takes these $K$ feature maps as inputs and produce the target image. 
We choose this keypoint-based representation because in our marble manipulation setting, the position of the marble and the depth of how much the marble is pressed into the gel capture the most relevant aspects of the state. 
The autoencoder network is trained self-supervisedly with L2 image reconstruction error, together with auxiliary losses that encourage sparse, non-redundant keypoints. 
In our experiments, we initially set the number of keypoints~$K=8$. 
We observed that all but one of the keypoints were inactive for all images, and the active keypoint location reliably matched the visible position of the marble on the \sensor{} images, while its intensity $i$ varied with the depth of the marble in the images -- the more the marble was pressed into the gel, the greater the intensity. 
We use the active keypoint as a compact representation of the raw \sensor{} image, and train one such keypoint autoencoder shared for both fingers. At the end of keypoint encoding, the state is represented by $s=[k_l, k_r, j]$, where $k_l$ and $k_r$ represent the keypoints from the left and right \sensor{}. This compact state representation is merely 14-dimensional, compared to the 64$\times$64 raw input images.
An overview of the learned model is shown in \fig{fig:mpc-diagram}.

We then train a neural network dynamics model $s' = f(s, a)$ on this state representation to predict the next state $s'$ conditional on the current state $s$ and action $a$. 
We sample $(s,a,s')$ tuples from the training set, and augment them in two ways: (i) insert several zero-action tuples of the form $(s,0,s)$ randomly, and (ii) perturb the RGB values and gamma-correct the images to increase the robustness of the model to changes in lighting.
Using the aforementioned state representation, the environment is fully observable. 
Therefore, we choose to use a simple multi-layer perceptron (MLP) over the more complicated variational recurrent neural network (VRNN) from~\citet{chung2015recurrent} as the dynamics model.
Having predicted the future keypoint representations $k_l$ and $k_r$, the future \sensor{} images can be reconstructed by passing these predictions through the decoder of the keypoint autoencoder trained above. Some examples are shown in \fig{fig:predictions}.
Our model (struct-NN), consisting of the structural autoencoder and the neural network dynamics model, is extremely lightweight since it only models the dynamics of the 14-D state representation. 
This makes fast inference affordable while using little memory compared to alternative visual predictive models for control, such as CDNA~\cite{Finn2016CDNA,Tian2019Manipulation}. 
This aspect of the struct-NN is critical to our ability to scale tactile-MPC~\cite{Tian2019Manipulation} to a multi-finger setting, as we will demonstrate in the next section.

\subsection{Model-based Control} 
After learning the dynamics model, we follow a similar procedure as~\citet{Tian2019Manipulation} and use model-predictive control (MPC) with the cross-entropy method (CEM) as the underlying optimization algorithm to perform in-hand marble manipulation.
However, with two \sensor{} images and 8 degrees of freedom (DOF), compared to their one tactile observation and 3 DOF, our search space for planning has much higher complexity.
In our setting, MPC with CEM requires predicting hundreds of thousands of possible future steps in the process of planning and executing one trajectory, which is prohibitively expensive when two \sensor{} images must be generated for each step. 
To overcome these difficulties, we plan directly in the 14-D state space instead of in the observation (image) space as in~\cite{Tian2019Manipulation}. 
Specifically, we first map our current image observations into the keypoint space using the keypoint encoder. Then, for generating prediction for each sequence of actions of length $T$, we only need to recursively apply the learned dynamics model to the 14-D state $s$ autoregressively $T$ times. 
Since the encoder network, the most computationally expensive part of the entire model, is only called once for each step of MPC (to map from images into keypoint space at the beginning), this optimization process becomes very inexpensive.

Given a goal \sensor{} image from one of the fingers, specifying a target position of the marble with respect to that finger, we first map it into the keypoint space as $k^g_l$ or $k^g_r$ (for the left or right finger). 
In our experiments, for simplicitly, we directly provide the target marble locations as the keypoint locations. 
During planning, the cost for each sequence of actions is the sum of Euclidean distance between the current position and the target position in $(x, y, i)$ coordinates. 
This encourages the planner to move the marble to the desired $(x, y)$ positions and also avoid dropping the marble or pressing it too hard.


\section{EXPERIMENTAL RESULTS}
\label{sec:result}

In addition to evaluate our design in terms of the quality of tactile images produced (\fig{fig:impressions}) and the robustness of the gel (\sec{sec:robustness}), we now evaluate the \sensor{} in the complex in-hand tactile manipulation task described in \sec{sec:approach}.

\begin{table}[t]
\centering
\caption{Comparison between Struct-NN and CDNA.}
\begin{tabular}{l|c|c}
\textbf{Performance} & Struct-NN~\citep{Minderer2019Unsupervised} & CDNA~\citep{Finn2016CDNA}  \\ \hline    
1 forward-backward pass  & \SI{4.3}{\milli\second} & \SI{6.8}{\milli\second} \\
1 forward pass & \SI{1.6}{\milli\second} & \SI{2.3}{\milli\second} \\
1 MPC step & \SI{1.4}{\second}& \SI{69}{\second} \\
\# of parameters & 1.2 M & 4 M \\
RMSE error (BAIR pushing) & 0.06023 & 0.01082 \\
RMSE error (Marble) & 0.00657 & 0.00028 \\

\end{tabular}
\label{tab:model}

\end{table}

\subsection{Video Predictive Model}
\begin{figure}[t]
    \centering
    \includegraphics[width=\linewidth]{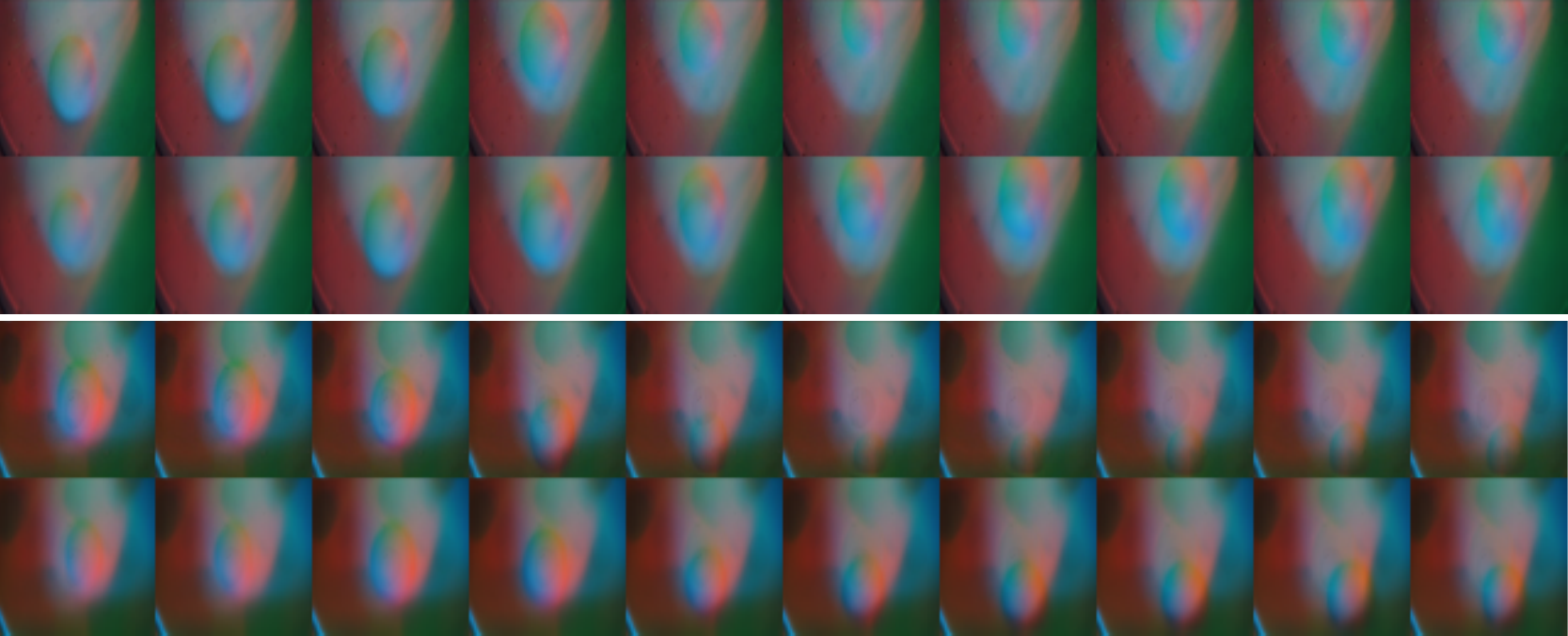}
    \caption{Sequences of trajectory predictions produced by our video-predictive model. The first and the third rows are ground truth images. The second and the fourth rows are image predicted and reconstructed by Struct-NN. The first 2 columns are context frames shown to the model, and the following 8 columns are predictions.}
    \label{fig:predictions}
\end{figure}

First, we evaluate the video predictive model alone. To validate our modeling choices, we measure the prediction error on a standard benchmark for video prediction, the BAIR robot pushing dataset~\cite{ebert2017self}, in addition to our \sensor{} tactile marble manipulation videos. 
In both datasets, we use 64 $\times$ 64 images and compare prediction performance with CDNA~\cite{Finn2016CDNA} used for tactile servoing in~\cite{Tian2019Manipulation} in terms of per-pixel root mean squared error (RMSE) on images in range $[0,1]$ as well as model sizes, training and inference time, and time for 1-step MPC. 
These results are shown in \tab{tab:model}.
Struct-NN produces qualitatively good predictions, as shown in \fig{fig:predictions}, but produces slightly higher RMSE than CDNA on BAIR pushing as well as our tactile marble manipulation videos. 
However, its primary advantage is its speed. 
In our multi-finger marble manipulation setup, MPC optimization is difficult and computationally demanding, requiring 250 particles with a planning horizon of 10 in each CEM iteration, for an average of 120 CEM iterations (about 0.3 million forward passes through the dynamics model) for a single MPC step. 
With the Struct-NN keypoint dynamics model, this step requires 1.4 seconds of computation. In comparison, CDNA would take 69 seconds for a single step, making it impractical to use for control.

\subsection{Manipulating Marbles}
\label{sec:servo_results}
\begin{figure}[t]
    \centering
    \includegraphics[width=.98\linewidth]{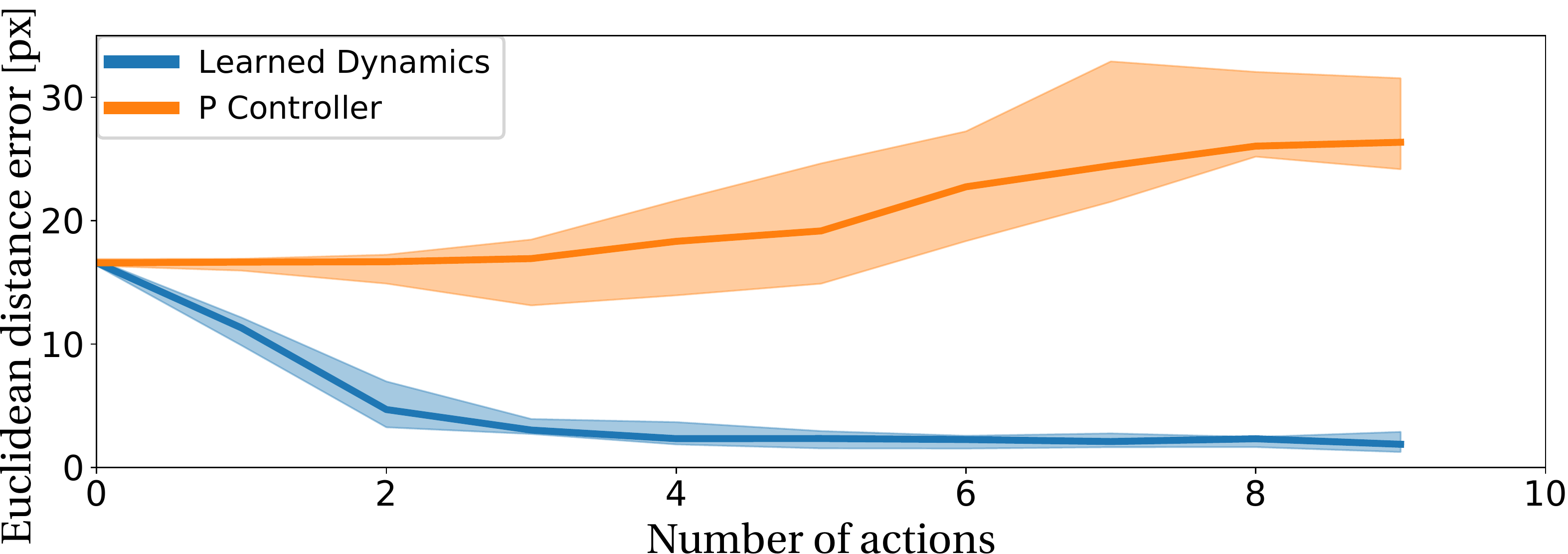}\\
    \includegraphics[width=.98\linewidth]{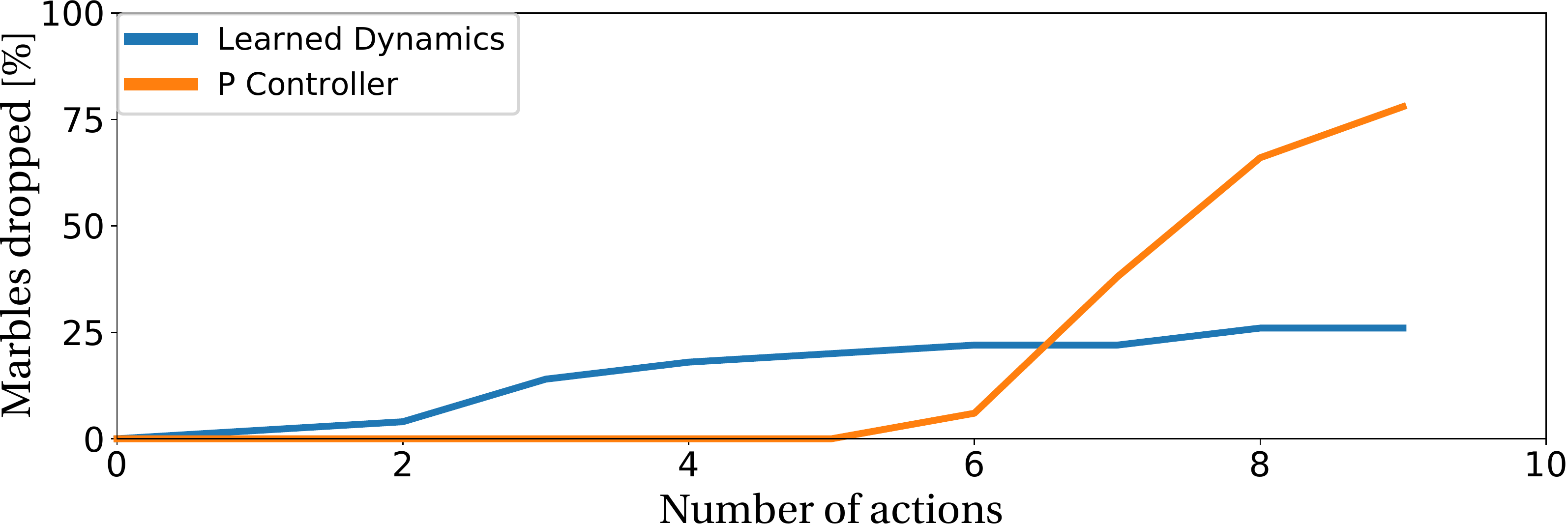}
    \caption{Results from real-world marble manipulation. (\textit{Top}) Euclidean distance (median and 68th percentile) to the desired goal during trajectory rollouts of MPC. The curve shows that our controller gets closer to the desired goal over time, while the hand-tuned P controller diverge on average. (\textit{Bottom}) Due to control noise, potential planning inaccuracies and the challenging nature of this task, the hand tends to drop marbles over time. }
    \label{fig:mpc_exp_result}
\end{figure}

We now evaluate the control task of manipulating a marble between two fingers.
This is a very challenging task because it requires controlling the slipping and rolling dynamics of the marble over the small and deformable \sensor{} surfaces under different pressure and joint positions, as well as maintaining enough force to hold the marble, but not too much to shoot the marble out of the hand.
In our experiments, after picking up the marble, we set the goal by setting the intensity $i$ to 1.0 and sampling $(x, y)$ in the keypoint space randomly and uniformly, under the constraint that it is at least 16 pixels away from the current marble position. 
We repeat each experiments 50 times to compute statistical performance.
One question that arise, is whether our MPC with learned non-linear dynamics model is necessary to control the marble once we learn the compact keypoints representation, or whether a simpler control scheme could be used.
To test this hyphothesis, we compared our approach against a simple linear proportional controller in keypoints space.
One challenge of comparing against the proportional controller is that the gains~$P$ consists of a 3$\times$8  matrix, which is multiplied against the 3-dimensional displacement vector between the current and the desired position in the keypoint $(x, y, i)$ space to produce the prescribed 8-D action. 
In our experiments, we manually tuned the gains based on human expertise and iterative trials.
However, compared to our MPC approach which is virtually parameters-free, this proved significantly more challenging.

\begin{figure}[t]
    \centering
    \includegraphics[width=\linewidth]{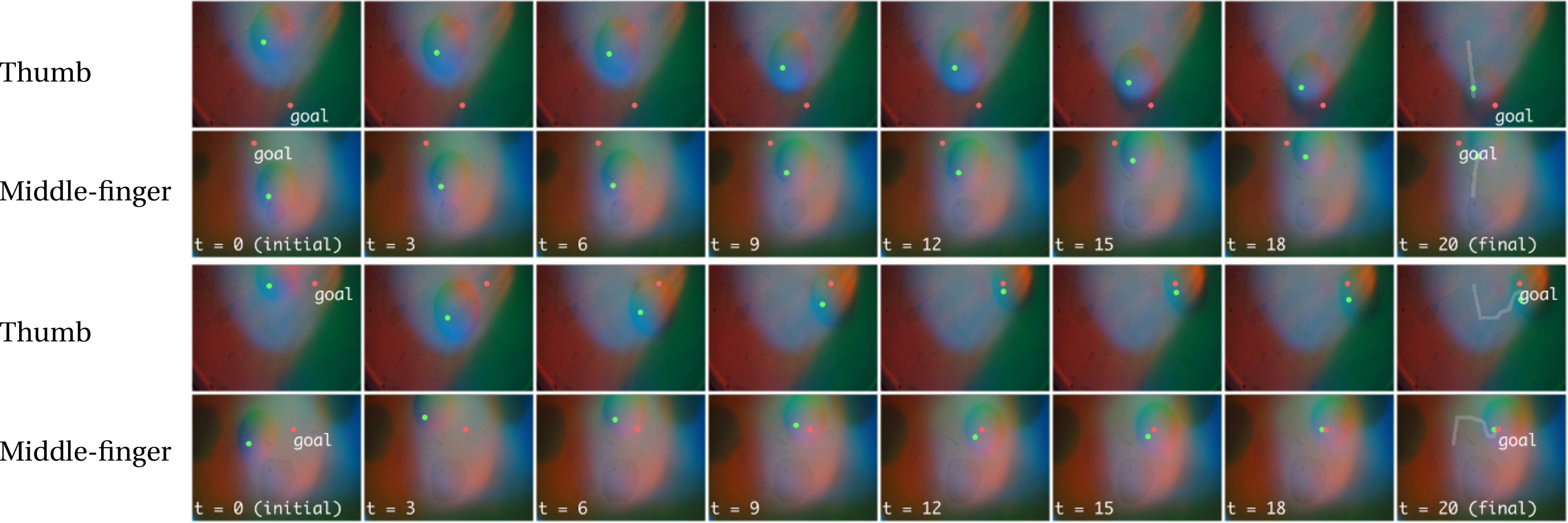}
    \caption{Examples of trajectories generated by MPC in tactile space. The goal position is marked by the red dot and the current keypoint position is represented by the green dot. In the last frame, we overlay the complete trajectory on top of the image. It can be seen how the MPC controller can move the marble to reach the goal quite accurately.}
    \label{fig:mpc_img_seq}
\end{figure}

The results of our evaluation are presented in \fig{fig:mpc_exp_result} where we plot the $(x, y)$ Euclidean distance to goal in pixels versus the number of actions performed. 
As \fig{fig:mpc_exp_result} (top) shows, the distance to the desired goal drops steadily over time for the learned dynamics, while it increases for the hand-tuned P controller. 
\fig{fig:mpc_img_seq} shows examples of trajectories from the learned model successfully rolling the marble to the desirable goal.
This result is in agreement with previous results in \citep{Tian2019Manipulation}, where learned models outperform simple hand-tuned controllers.
However, about 25\% of trials result in the marble dropping before it can be fully manipulated to the goal, as shown in \fig{fig:mpc_exp_result} (bottom). 
Aside from the challenging nature of this task, we believe that this is partly due to planning inaccuracies and actuation noise in the hand joints. 
We hypothesize that improving the low level controller and collecting more data for improving the learned model will help in decreasing the number of marbles dropped over time and further improve performance.
As for the poor performance of the linear P controller, we suspect one reason might be the non-linearity of the dynamics: the end effectors (normal vectors of \sensor{} surfaces) are non-trivial trigonometric functions of the joint displacement commands of each finger we are controlling, but additionally, the surfaces of contact at the elastomeric gel of the \sensor{}s, are curved and deformable. 
A fixed P matrix can only be optimal in some of the operating regions but not all of them, especially at the boundary of the robot configuration space where some of the joint angles reach the limits of the actuator. 
Even if it is possible to find a sub-optimal P matrix that works for most of the time, it is still non trivial.

The results obtained on this marble manipulation task validates both our key contributions.
First, it shows that \sensor{} provides high resolution tactile sensing capable of such fine-grained and challenging multi-fingered in-hand manipulation tasks. 
Second, it shows that our solution to scaling up tactile MPC using Struct-NN can successfully handle the complexity of this task.


\section{CONCLUSION}
\label{sec:conclusion}

	Tactile sensing is an important component towards human-level manipulation skill for robots.
In this paper, we present a new compact tactile sensor -- \sensor{} -- which provides rich, high-resolution tactile readings.
In addition, \sensor{} provides significant improvements across many other valuable metrics: reliability, component availability, ease of assembly, and manufacturing cost.
We demonstrate the capabilities of this new sensor by tackling a challenging fine motor control task: in-hand marble manipulation.
Building on advances in deep model predictive control, we learn to manipulate glass marbles from raw tactile inputs towards desired target positions. 
We believe that \sensor{} is a step forward in the design of versatile tactile sensors that can be mass-produced and widely adopted in the robotic community towards reaching human-level manipulation skills.
For this purpose, we open-source the design and manufacturing process of \sensor{} at \repo{}.
Future work should aim at further miniaturizing the form factor of the sensor, and designing sensors with curved, omni-directional sensing fields.





\section*{ACKNOWLEDGMENT}
We thank Wenzhen Yuan and Ted Adelson for insightful discussions; Nolan Black, Spencer Burns, Allan Smith, Jake Khatha, Louks Hendricks and Area 404 for supporting the manufacturing of the sensor; GelSight Inc. for providing a gel for comparison.


\bibliographystyle{IEEEtran}
\bibliography{paper}

\end{document}